\documentclass[10pt,twocolumn,letterpaper]{article}

\usepackage[pagenumbers]{cvpr} % To force page numbers, e.g. for an arXiv version

\definecolor{cvprblue}{rgb}{0.21,0.49,0.74}
\usepackage[pagebackref,breaklinks,colorlinks,allcolors=cvprblue]{hyperref}
\usepackage{graphicx} % Required for inserting images
\usepackage[acronym]{glossaries}
\usepackage{bm}
\usepackage{siunitx} % For scientific notation
\usepackage[accsupp]{axessibility}  % Improves PDF readability for those with disabilities.

\newacronym{InstructPix2Pix}{InstPix2Pix}{instructpix2pix}

\def\paperID{XAI4CV-14} % *** Enter the Paper ID here
\def\confName{CVPRW}
\def\confYear{2025}

\title{X-Edit: Detecting and Localizing Edits in Images Altered by Text-Guided Diffusion Models}

\author{Valentina Bazyleva\\
Reality Defender\\
{\tt\small bazvalya@gmail.com}
\and
Nicol\`o Bonettini\\
Reality Defender\\
{\tt\small nicolo@realitydefender.com}
\and
Gaurav Bharaj\\
Reality Defender\\
{\tt\small gaurav@realitydefender.com}
}

\begin{document}
\maketitle
\begin{abstract}
    Text-guided diffusion models have significantly advanced image editing, enabling highly realistic and local modifications based on textual prompts. While these developments expand creative possibilities, their malicious use poses substantial challenges for detection of such subtle deepfake edits.
    To this end, we introduce Explain Edit (X-Edit), a novel method for localizing diffusion-based edits in images. To localize the edits for an image, we invert the image using a pretrained diffusion model, then use these inverted features as input to a segmentation network that explicitly predicts the edited ``masked'' regions via channel and spatial attention. 
    Further, we finetune the model using a combined segmentation and relevance loss. The segmentation loss ensures accurate mask prediction by balancing pixel-wise errors and perceptual similarity, while the relevance loss guides the model to focus on low-frequency regions and mitigate high-frequency artifacts, enhancing the localization of subtle edits.
    To the best of our knowledge, we are the first to address and model the problem of localizing diffusion-based modified regions in images. We additionally contribute a new dataset of paired original and edited images addressing the current lack of resources for this task.
    Experimental results demonstrate that X-Edit accurately localizes edits in images altered by text-guided diffusion models, outperforming baselines in PSNR and SSIM metrics. This highlights X-Edit's potential as a robust forensic tool for detecting and pinpointing manipulations introduced by advanced image editing techniques.
\end{abstract}

% \gb{media authentication and digital forensics}, where
    % X-Edit combines inversion-based feature extraction from a pretrained diffusion model with a segmentation network that's enhanced via channel and spatial attention to localize the edits. 
% Later, to explicitly balance high and low frequency image features introduced during the inversion of edited regions, we incorporates additional a relevance losses.
    
\section{Introduction}
\label{sec_introction}

The rapid advancements in generative models have revolutionized computer vision, particularly with the rise of diffusion models as a compelling alternative to GANs \cite{goodfellow2014generative}. Unlike GANs, diffusion models offer greater stability and are capable of generating high-quality, diverse images while avoiding issues such as mode collapse \cite{ho2020denoising, sohl2015deep, dhariwal2021diffusion, song2020denoising}. Their strength is in mapping images across domains while maintaining essential properties, which makes them very useful for tasks such as Image-to-Image (I2I) translation \cite{pang2021image, isola2017image}. One specific application is text-guided image editing (TGIE) \cite{huang2024diffusion}, which allows for localized modifications to images based on textual prompts, seamlessly altering content while maintaining photorealism. While these features expand artistic possibilities, they also pose serious problems for media authentication and digital forensics since it gets increasingly challenging to identify and interpret these kinds of modifications.

\begin{figure}[t]
    \centering
    \includegraphics[width=\columnwidth]{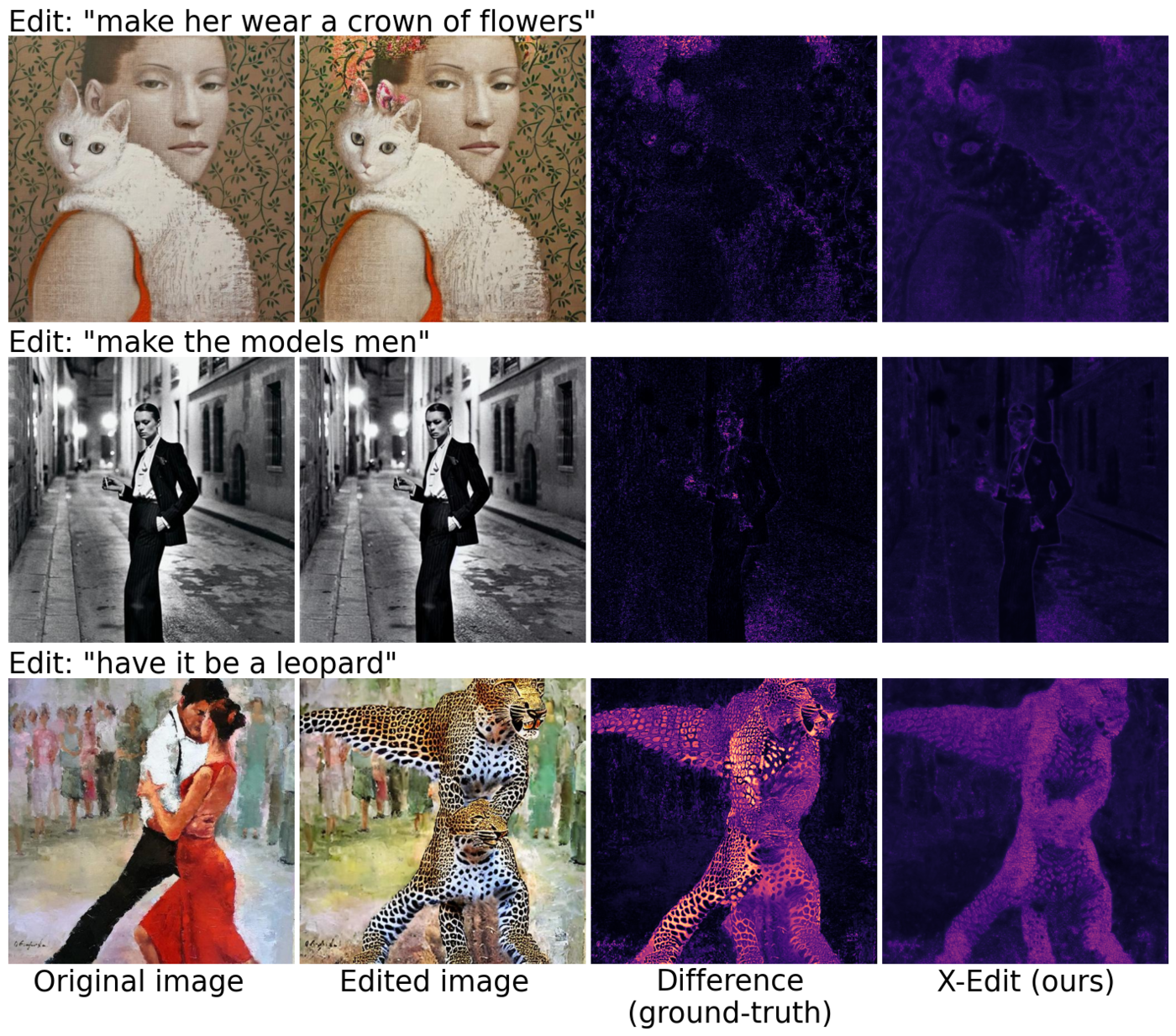}
    \caption{\textbf{Examples of our X-Edit localization method.} X-Edit is able to localize text-guided image edits disregarding  areas not affected by the edit. 
    % \TODO{change X-Edit localization in label}
    }
    \label{fig:teaser}
\end{figure}

Traditional methods for detecting AI-generated content have focused largely on binary classification, aiming to distinguish between real and synthetic images \cite{yu2019attributing, wang2020cnn, frank2020leveraging}. However, TGIE presents a more complex challenge: it is not enough to detect whether an image has been manipulated; it is equally important to pinpoint the exact regions where the edits have occurred. In fields like digital forensics, accurately localizing these altered areas is crucial for deeper analysis and validation.

\begin{figure}[t]
    \centering
    \includegraphics[width=\columnwidth]{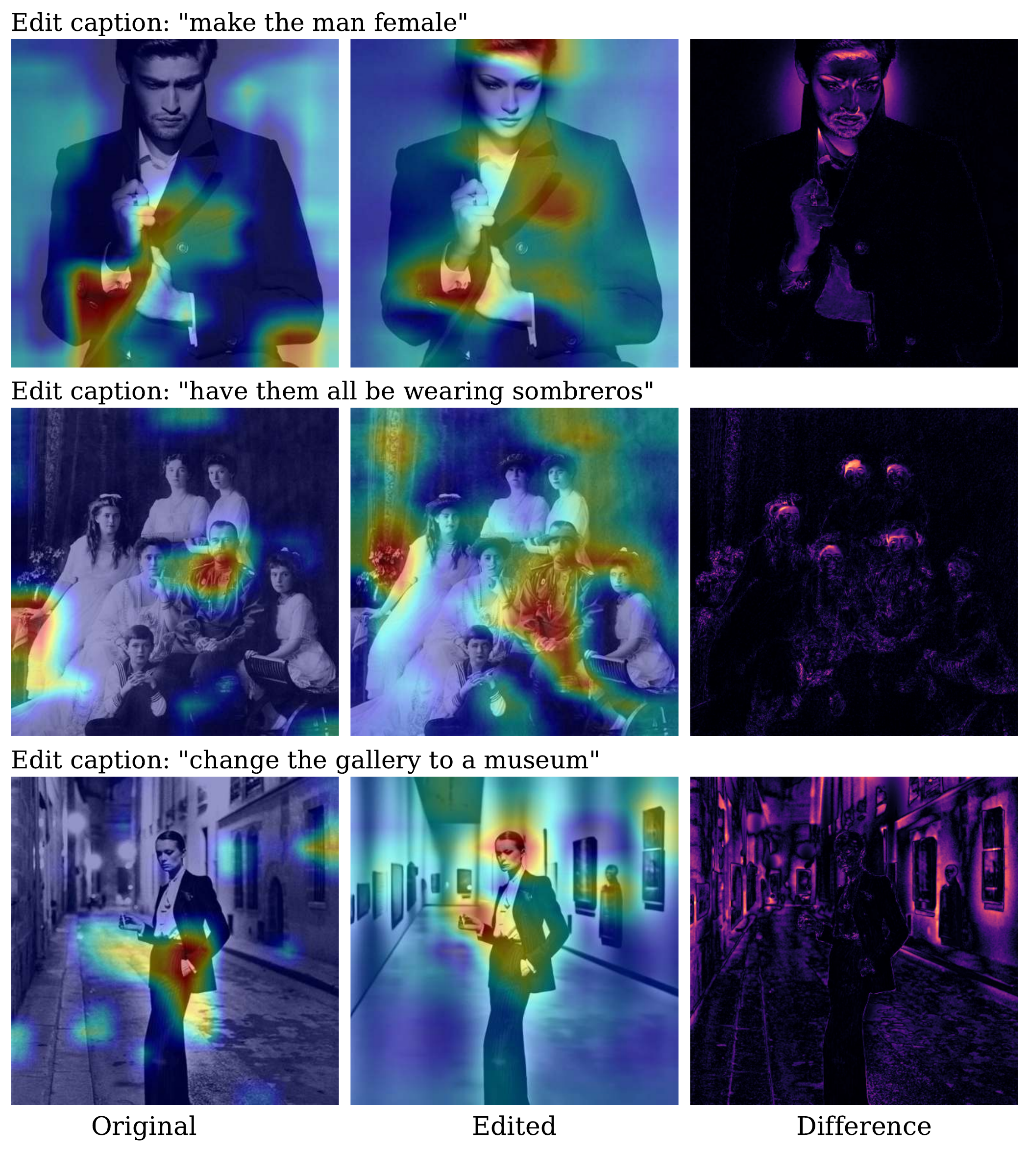}
    \caption{\textbf{Grad-CAM visualizations for original and edited images.} The results highlight that the coarse relevance maps produced by EfficientNet with high detection accuracy (99.93\%) and precision (99.92\%) do not accurately localize specific edit regions.}
    \label{fig:grad-cam}
\end{figure}

Explainable artificial intelligence (XAI) techniques, such as Grad-CAM, have been proposed to provide insights into model decision-making processes \cite{adadi2018peeking, selvaraju2017grad}. 
\cref{fig:grad-cam} demonstrates how Grad-CAM produces broad relevance maps that struggle to precisely localize edits. For instance, gender alteration (row 1) and the addition of sombreros (row 2) result in diffuse activations rather than alignment with the specific edited areas. This imprecision limits the utility of conventional relevance mapping in forensic tasks, where accurate edit localization is essential.
% While useful for classification, these methods often yield coarse relevance maps that do not align with edited areas, making them insufficient for accurately localizing these regions and thus limiting their utility in forensic analysis. 

Upon literature review, we notice a lack of datasets to tackle the TGIE localization task. In this work, we aim to fill that gap by providing our dataset consisting of a total of \num{167026} pairs of original and edited images, which were altered with diffusion-based TGIE methods like InstructPix2Pix \cite{instructpix2pix_2023}.
We also introduce a new approach that not only identifies image alteration but also precisely localizes the modified regions, thereby addressing the shortcomings of previous methods.
Our approach employs our new dataset to train a model that can predict the masks of the edited regions. We draw on features extracted from the inversion of images through a pretrained Stable Diffusion model, similar to the FakeInversion approach \cite{cazenavette2024fakeinversion}, which captures rich information about discrepancies introduced during editing. 
FakeInversion lacks the ability to localize the edits, and we address this limitation.
We employ a model with U-Net architecture and attention mechanism~\cite{ronneberger2015u, woo2018cbam} that is first trained on these inversion features and then finetuned with a relevance-based loss function~\cite{chefer2022optimizing} that encourages the model’s relevance maps to concentrate on the altered regions while maintaining overall classification performance. With this method the model is encouraged to prioritize the identification of modified areas. We demonstrate our method's efficacy localizing edits introduced by text-guided diffusion models by validating it on a large paired dataset of original and edited images. To summarize, we contribute the following: 
\begin{enumerate}
    % new problem
    \item We introduce new problem to detect local edits in images due to diffusion-based text-guided image editing (TGIE) methods. 
    %  new dataset for the problem
    \item We present a new paired dataset of original and edited images, providing a valuable resource for TGIE detection and localization research.
    % new method for the problem
    \item We propose a novel method that leverages inversion discrepancies from pretrained diffusion models and uses a specialized segmentation network with attention mechanisms and a composite loss function, enabling precise localization of edited regions in images.
    % \item \TODO{We propose a method that combines FakeInversion-like features with an XAI-driven model to achieve precise localization of edited regions.}
    %  "SOTA" 
    \item Our method outperforms several baselines in terms of PSNR and SSIM, with balanced precision and recall.
\end{enumerate}

\section{Related Work}
\label{sec_related_work}

% This section situates the paper within the growing body of research on the detection of imagine forgeries, more especially on the detection of images altered with diffusion model-based editing methods.

\subsection{Image-to-Image Translation in the Context of Diffusion Models}

I2I translation involves transforming an image from one domain to another while preserving its core structures, such as object shape and layout \cite{pang2021image}. Traditionally, GANs have been influential in this domain, addressing tasks like inpainting, restoration, and colorization \cite{saharia2022palette, saxena2021comparison}. Recently, diffusion models have emerged as a more effective alternative, particularly for TGIE, where images are modified based on textual prompts \cite{tumanyan2023plug}. These models enable localized changes that semantically align with target instructions.

A recent comprehensive survey \cite{huang2024diffusion} classifies diffusion-based image editing methods into three categories: (1) \textit{training and finetuning free approaches}, (2) \textit{training-based approaches},  and (3) \textit{testing-time finetuning approaches}. Our work focuses on methods from the first two groups. \textit{Training and finetuning free approaches} leverage inherent features of diffusion models to achieve editing without additional training. Attention mechanisms in models like Stable Diffusion's U-Net \cite{rombach2022high} are exploited to control image alterations. Techniques such as Prompt-to-Prompt (P2P) \cite{hertz2022prompt} replace attention maps to maintain consistency between edited and original images. \mbox{MasaCtrl~\cite{cao2023masactrl}} modifies self-attention by replacing key and value features, enhancing action editing, while Plug-and-Play (PnP) \cite{tumanyan2023plug} adjusts query and key features for finer control. Free-Prompt Editing (FPE) \cite{liu2024towards} selectively alters attention maps to preserve geometric details. \textit{Training-based approaches} involve finetuning models for specific editing tasks, offering greater precision at the cost of additional resources. InstructPix2Pix \cite{instructpix2pix_2023} and MLLM-Guided Image Editing (MGIE) \cite{fu2023guiding} combine the flexibility of prompt-based methods like P2P with supervised training to achieve greater precision.

Inversion-based techniques are also critical for controlled I2I editing. Methods like DDIM and DDPM inversion recover latent representations for accurate reconstructions \cite{song2020denoising, huberman2024edit}. Null-text Inversion \cite{mokady2023null} aligns reconstruction paths using optimized embeddings, while Negative-Prompt Inversion \cite{miyake2023negative, han2024proxedit} reduces computational costs. DICE \cite{he2024dice} extends these approaches to discrete diffusion models, enabling finegrained editing without predefined masks or attention manipulations.

\subsection{Detectors of Diffusion-based Generated Images}

Early detection efforts for AI-generated content focused on GANs, utilizing frequency domain analysis \cite{zhang2019detecting, marra2019gans} and deep learning classifiers \cite{wang2019fakespotter, wang2020cnn} for the detection. With the rise of diffusion models, these methods adapted accordingly. Studies have shown that GAN detectors perform poorly on diffusion-generated images unless properly retrained \cite{ricker2022towards, corvi2023detection}.

Specific detection methods for diffusion models include DIRE \cite{wang2023dire}, which detects discrepancies between original and reconstructed images, and FakeInversion \cite{cazenavette2024fakeinversion}, which uses inverted features from Stable Diffusion. These methods are inspired by earlier findings that CLIP embeddings \cite{radford2021learning} can predict image authenticity \cite{ojha2023towards, sha2023fake}. Enhancements like DistilDIRE \cite{lim2024distildire} improve efficiency, while MultiLID \cite{lorenz2023detecting} uses intrinsic dimensionality analysis. Hybrid models combining attention-guided feature extraction with vision transformers further enhance detection accuracy \cite{xu2023exposing}.

A comprehensive review of the difficulties and developments in the identification of diffusion-generated images is provided in \cite{lee2024tug}, highlighting the necessity of ongoing innovation in forensic methods to keep pace with the rapid advancement of generative models.

\begin{figure*}[htbp]
    \centering
    \includegraphics[width=\textwidth]{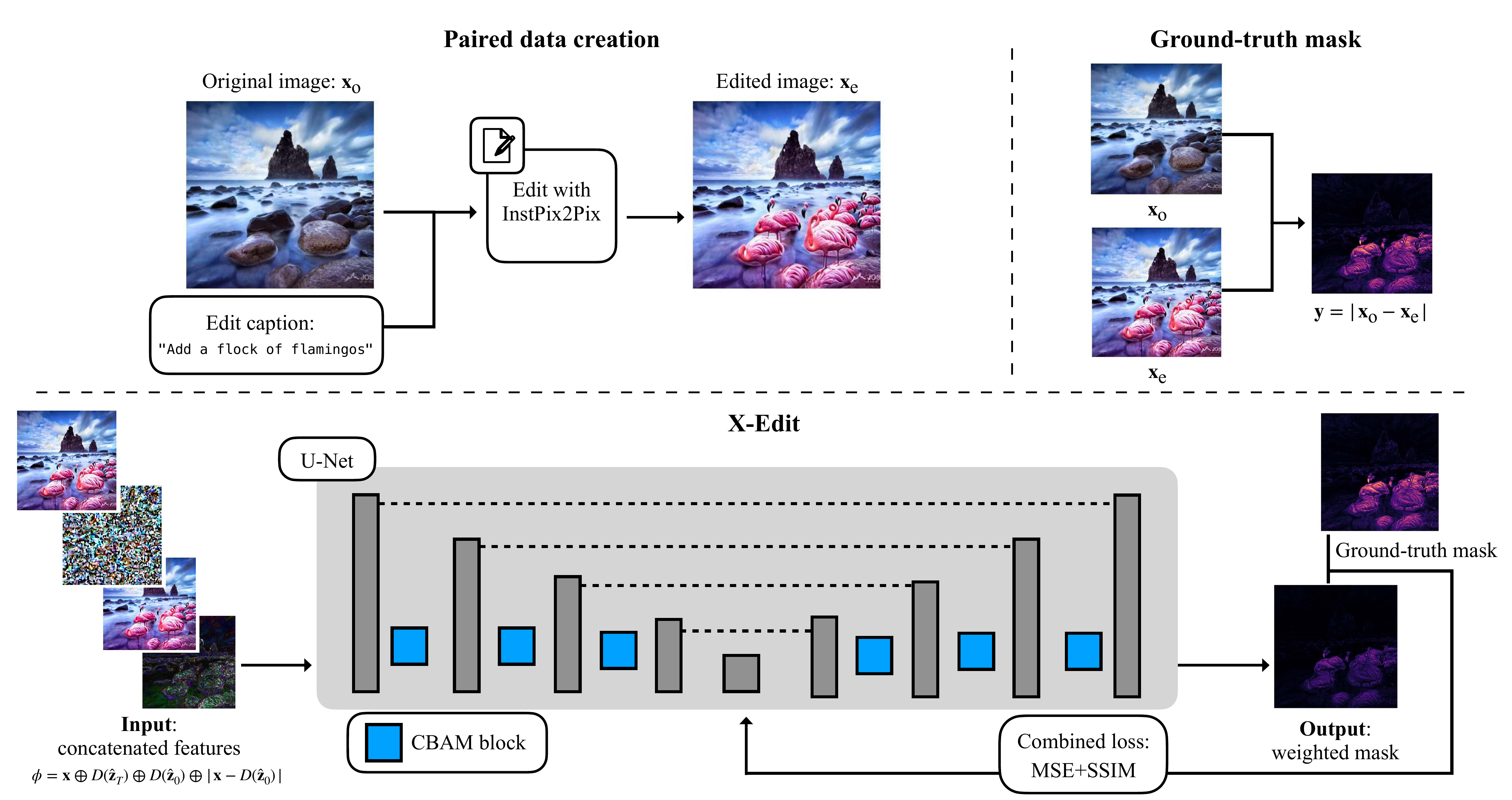}
    \caption{\textbf{Overview of the X-Edit framework.} The pipeline begins with paired data creation, where each image is modified with InstructPix2Pix based on one specified edit caption to produce an edited image. A ground-truth mask is generated by computing the absolute difference between original and corresponding edited images. 
    % This mask is used as a target for training. 
    In the X-Edit, input features, including concatenated image, decoded noise, reconstructed image, and residual differences, are processed by a U-Net architecture with integrated CBAM blocks for enhanced attention. The model is optimized using a combined loss to generate an output mask that reconstructs the ground-truth mask, highlighting edited regions.}
    \label{fig:xedit_pipeline}
\end{figure*}

\subsection{Detection and XAI Methods for Images Edited with Image-to-Image Diffusion Models}

Detecting images edited using I2I diffusion models poses challenges due to subtle edits and preserved original content. Traditional methods like error level analysis and artifact detection \cite{farid2006exposing} are less effective against advanced generative models that produce minimal artifacts. With the rise of GANs and diffusion models, researchers developed methods to detect manipulations by examining model-specific artifacts \cite{marra2018detection}.

XAI tools are crucial for interpreting model decisions in image editing detection \cite{arrieta2020explainable}. Grad-CAM \cite{selvaraju2017grad} produces localization maps highlighting important regions influencing model predictions. Some methods learn manipulation fingerprints using constrained convolutional layers \cite{bayar2016deep}. Hybrid architectures combining spatial and temporal features have been proposed for detecting and localizing image forgeries \cite{bappy2019hybrid}. Metrics like VIEScore \cite{ku2023viescore} provide explainable assessments for conditional image synthesis, while approaches like X-IQE \cite{chen2023x} use language models to generate textual explanations of image quality and alignment.

Parallel research has focused on making the diffusion process interpretable.  Visualization tools have been developed to illustrate image generation at each denoising step \cite{park2024explaining}, while other studies have examined the trustworthiness of text-to-image diffusion models \cite{zhang2024trustworthy}, emphasizing the need for interpretability.

Despite advancements, a gap remains in techniques that detect and explain text-guided diffusion edits. Integrated tools are essential for media authentication, where understanding how an image was manipulated is as crucial as detecting whether it was edited.

\section{Method}
\label{sec_method}

In this section, we present a detailed description of how our dataset is created, as well as \textbf{X-Edit}, a method for detecting and localizing edits in images altered by diffusion-based TGIE methods. X-Edit leverages discrepancies arising during the inversion of a pretrained diffusion model to identify and localize modifications within an image.

\paragraph{Dataset} We construct a dataset consisting of \num{167026} pairs of original and edited images, resulting in a total of \num{334052} images. The edited images are generated using InstructPix2Pix \cite{instructpix2pix_2023}, a TGIE method. The dataset is created by utilizing image URLs from the LAION-Aestetics V2 6.5+ dataset \cite{schuhmann2022laion}, accompanied by captions and edit instructions generated by the InstructPix2Pix authors. That is, using a finetuned version of GPT-3 \cite{NEURIPS2020_1457c0d6}, they generated text triplets: a caption describing the original image, an edit instruction, and a caption describing the image after the edit. This method ensured consistency in describing a diverse range of edits across the original and edited images. We filter the downloaded images from LAION to include only those with a resolution of at least $400\times400$ pixels, ensuring editing quality. Using the generated text instructions, we apply  InstructPix2Pix to produce the edited versions of the images. Dataset will be released at \url{https://github.com/Reality-Defender/X-Edit-CVPRW25}.

It is important to note that while InstructPix2Pix authors generated different versions of original images using Stable Diffusion and then applied edits with the instructions, we directly applied the edits to the original images from the LAION dataset. This approach preserves the authenticity of the original content and provides a more realistic dataset for training and evaluating edit localization methods.

\paragraph{Feature Extraction} X-Edit builds upon the FakeInversion method \cite{cazenavette2024fakeinversion} by extracting features that capture inconsistencies between an image and its reconstruction through the inversion of Stable Diffusion. Given an input image $\mathbf{x}$, we encode it into the latent space using the VAE encoder:
\begin{equation}
    \mathbf{z}_0 = \text{VAE}_{\text{enc}}(\mathbf{x}).
\end{equation}
We perform DDIM inversion \cite{song2020denoising} to estimate the initial noise map $\hat{\mathbf{z}}_T$ that would generate $\mathbf{z}_0$ in the forward diffusion process, conditioned on the CLIP text embedding $\mathbf{c}$ of a predicted caption $c$:
\begin{equation}
 \hat{\mathbf{z}}_T = \text{DDIMInversion}(\mathbf{z}_0, \mathbf{c}).
\end{equation}
We reconstruct the latent image $\hat{\mathbf{z}}_0$ from $\hat{\mathbf{z}}_T$ by conditional reverse diffusion process and decode both $\hat{\mathbf{z}}_T$ and $\hat{\mathbf{z}}_0$ back to the image space:
\begin{equation}
    D(\hat{\mathbf{z}}_T) = \text{VAE}_{\text{dec}}(\hat{\mathbf{z}}_T), \quad D(\hat{\mathbf{z}}_0) = \text{VAE}_{\text{dec}}(\hat{\mathbf{z}}_0).
\end{equation}
As shown in~\cite{cazenavette2024fakeinversion}, the inversion introduces discretization errors $\delta$, causing discrepancies between $\hat{\mathbf{z}}_0$ and $\mathbf{z}_0$ that are informative for detecting edits, as they are more pronounced in altered regions. 
% The log-likelihood relationship is given by \cite{cazenavette2024fakeinversion}:
Then, the log-likelihood of the data given the underlying model can be approximated in the first order by:
\begin{equation}
    \log p(\mathbf{z}_0) \propto \log p_z(\mathbf{z}_T) - \frac{\langle \delta, \hat{\mathbf{z}}_0 - \mathbf{z}_0 \rangle}{\lVert\delta\rVert^2} .
\end{equation}
This relationship indicates that significant discrepancies between the reconstructed and original latents, especially in regions with edits, lead to a lower log-likelihood. Therefore, analyzing these discrepancies enables us to detect and localize edits within images.

\paragraph{Edit Localization with X-Edit} 
To localize edits, X-Edit uses a composite input, which is constructed by concatenating the image $\mathbf{x}$, the decoded noise map $D(\hat{\mathbf{z}}_T)$, the reconstructed image $D(\hat{\mathbf{z}}_0)$, and the reconstruction residual $|\mathbf{x} - D(\hat{\mathbf{z}}_0)|$:
\begin{equation}
    \bm{\phi} = \mathbf{x} \oplus D(\hat{\mathbf{z}}_T) \oplus D(\hat{\mathbf{z}}_0) \oplus |\mathbf{x} - D(\hat{\mathbf{z}}_0)|,
\label{eq:phi}
\end{equation}
where $\oplus$ represents the concatenation operation along the channel axis.
This input aims to encapsulate discrepancies that are more prominent in edited regions. 
%
% We train a segmentation model to predict a mask $\hat{\mathbf{y}}$, indicating the likelihood of each pixel belonging to an edited area, using ground truth masks $\mathbf{y}$ computed as:
% \begin{equation}
% \label{eq:gt}
%     \mathbf{y} = | \mathbf{x}_{\text{o}} - \mathbf{x}_{\text{e}} |,
% \end{equation}
% where $\mathbf{x}_{\text{o}}$ and $\mathbf{x}_{\text{e}}$ represents the original and edited images respectively. 
%
After defining a ground-truth mask $\mathbf{y}$ as:
\begin{equation}
\label{eq:gt}
    \mathbf{y} = | \mathbf{x}_{\text{o}} - \mathbf{x}_{\text{e}} |,
\end{equation}
where $\mathbf{x}_{\text{o}}$ and $\mathbf{x}_{\text{e}}$ represents the original and edited images respectively, we train a segmentation model to predict a mask $\hat{\mathbf{y}}$, indicating the likelihood of each pixel belonging to an edited area.

By learning the association between inversion-induced discrepancies and edited regions, the model effectively localizes edits. Conditioning the inversion on the predicted caption $c$ amplifies discrepancies when the image content deviates from expectations due to edits. Thus, X-Edit extends inversion-based feature extraction to the task of edit localization by combining discrepancies captured during inversion with a segmentation model. This approach provides a robust method for detecting and precisely localizing edits introduced by diffusion-based TGIE methods.

\paragraph{Finetuning Procedure} To further enhance the model's ability to localize edits, we adopted a finetuning strategy inspired by the RobustViT method proposed in \cite{chefer2022optimizing}. The goal of this finetuning procedure was to direct the model's attention toward the edited regions while reducing the focus on irrelevant background features. To achieve this, we incorporated an additional relevance loss, encouraging the model's relevance map $R(\mathbf{x})$ to resemble the ground truth mask. Relevance maps were generated using integrated gradients \cite{sundararajan2017axiomatic}, which quantify the contribution of each input pixel to the model's prediction. 
% To guide the model's attention more effectively, we computed high-frequency $H(\mathbf{x})$ and low-frequency $L(\mathbf{x})$ maps using the gradient magnitude of the image:

While the original RobustViT method focuses on foreground and background regions to guide the model's attention, we modify this approach by distinguishing between high-frequency and low-frequency components in the image. 
Specifically, we compute high-frequency $H(\mathbf{x})$ and low-frequency $L(\mathbf{x})$ components using the Sobel filter to calculate the gradient magnitude of the image, which emphasizes edges and transitions. The Sobel filter computes gradients along the horizontal and vertical directions, capturing changes in pixel intensities.
%
% \TODO{
% Specifically, we computed high-frequency $H(\mathbf{x})$ and low-frequency $L(\mathbf{x})$ maps using 
% %
% the gradient magnitude of the image:
% \begin{equation} 
%     G(\mathbf{x}) = \sqrt{\left( \frac{\partial \mathbf{x}}{\partial u} \right)^2 + \left( \frac{\partial \mathbf{x}}{\partial v} \right)^2}, 
% \end{equation}
%  where $\frac{\partial \mathbf{x}}{\partial u}$ and $\frac{\partial \mathbf{x}}{\partial v}$ represent the gradients of the image $\mathbf{x}$ along the horizontal ($u$) and vertical ($v$) directions, respectively. 
%  %
 
%  % We then normalize $G(\mathbf{x})$ to obtain $G_{\text{norm}}(\mathbf{x})$.
%  We then normalize $G(\mathbf{x})$ to obtain $G_{\text{norm}}(\mathbf{x})$:
%  \begin{equation} 
%  G_{\text{norm}}(\mathbf{x}) = \frac{G(\mathbf{x}) - \min(G(\mathbf{x}))}{\max(G(\mathbf{x})) - \min(G(\mathbf{x})) + \epsilon}, 
%  \end{equation} 
%  where $\epsilon$ is a small constant added to prevent division by zero, and define the high-frequency and low-frequency components as:
% \begin{equation} 
%     H(\mathbf{x}) = G_{\text{norm}}(\mathbf{x}), \quad L(\mathbf{x}) = 1 - G_{\text{norm}}(\mathbf{x}).
% \end{equation}
% }
The relevance loss $\mathcal{L}_{\text{R}}$ is designed to penalize the model for assigning high relevance to edges while encouraging higher relevance in flat regions. We achieve this with:
\begin{equation} 
    \begin{split}
        \mathcal{L}_{\text{R}} = &\; \lambda_{\text{flat}} \cdot \text{MSE}(R(\mathbf{x}) \odot L(\mathbf{x}), 1) \\
        &+ \lambda_{\text{edge}} \cdot \text{MSE}(R(\mathbf{x}) \odot H(\mathbf{x}), 0),
    \end{split}
\label{eq:relevance_loss}
\end{equation}
where $\lambda_{\text{flat}}$ and $\lambda_{\text{edge}}$ are weighting coefficients, and $\odot$ is the Hadamard product. 
We also introduce the segmentation loss $\mathcal{L}_{\text{S}}$, integrating Mean Squared Error (MSE) and Structural Similarity Index Measure (SSIM) \cite{wang2004image} losses:
\begin{equation} 
    \mathcal{L}_{\text{S}} = \text{MSE}(\hat{\mathbf{y}}, \mathbf{y}) + \alpha \cdot [1 - \text{SSIM}(\hat{\mathbf{y}}, \mathbf{y})],
\end{equation}
where $\hat{\mathbf{y}}$ represents the model's predicted mask, $\mathbf{y}$ is the ground truth mask, and $\alpha$ is a weighting coefficient used to adjust the balance between pixel-wise error and perceptual similarity.
Our total loss function is a combination of the segmentation loss and relevance loss, ensuring that the model focuses on both accurate segmentation and attention to the correct regions:
\begin{equation} 
    \mathcal{L}_{\text{total}} = \lambda_{\text{R}} \cdot \mathcal{L}_{\text{R}} + \lambda_{\text{S}} \cdot \mathcal{L}_{\text{S}},
\end{equation}
where $\lambda_{\text{R}}$ and $\lambda_{\text{S}}$ are weighting coefficients.
% and $\mathcal{L}_{\text{S}}$ is 

\section{Experiments}
\label{sec_experiments}

% This section details the implementation specifics, including model architectures, inputs, baselines, evaluation metrics, and optimization.

\paragraph{Input Data Format} 
We experiment with three different input data format configuration. In the first configuration, we build $\bm{\phi}$ as defined in \cref{eq:phi} using the original image, the inverted image $D(\hat{\mathbf{z}}_T)$, the reconstructed image $D(\hat{\mathbf{z}}_0)$, and the residual error $(\mathbf{x} - D(\hat{\mathbf{z}}_0))$, each with 3 color channels (RGB), resulting in a total of 12 channels. The second configuration, $\bm{\phi}_{\mathrm{FI}}$, is defined as the input feature used in FakeInversion paper~\cite{cazenavette2024fakeinversion}. It simplifies the input to 3 grayscale channels representing the original image, inverted image, and reconstructed image. 
The third configuration is simply the input image $\mathbf{x}$.
Before generating the FakeInversion features, we use xGen-MM (BLIP-3)~\cite{xue2024xgen} to generate captions for the original and edited images, ensuring consistency between the text and image content. Additionally, we apply random augmentations such as horizontal flips, cropping, gaussian blur, and coarse dropout to both images, ensuring that the ground truth masks remained accurate post-augmentation.
For edited images, we construct the ground truth masks by computing the absolute difference between the edited image and the original image, as in \cref{eq:gt}. These masks highlight the edited regions, with non-zero values where modifications have occurred, and serve as the target labels for the model to predict. For the original images, the ground truth masks are zero tensors, as no modifications exist.

\begin{table}[t]
    \centering
    \resizebox{\columnwidth}{!}{%
    \begin{tabular}{l c c c}
        \toprule
        \textbf{Model} & \textbf{Input} & \textbf{PSNR} & \textbf{SSIM} \\
        \midrule

        % SAM~\cite{kirillov2023segment} & $\bm{\phi} = \mathbf{x}$ & \num{4.489e-3} & 23.478 & 0.506 \\
        SAM~\cite{kirillov2023segment} & $\mathbf{x}$ & 23.478 & 0.506 \\
        
        % SegFormer~\cite{xie2021segformer} & $\bm{\phi} = \mathbf{x}$ & 0.006 & 22.145 & 0.296 \\
        
        % U-Net & $\bm{\phi} = \mathbf{x}$ & \num{3.410e-3} & 24.672 & 0.902 \\
        U-Net & $\mathbf{x}$ & 24.672 & 0.902 \\
        U-Net & $\bm{\phi}_{\mathrm{FI}}$ & 24.785 & 0.919 \\

        ViT-B \cite{dosovitskiy2020image}  &  $\bm{\phi}_{\mathrm{FI}}$ & 24.772 & 0.875 \\ 
        SegFormer \cite{xie2021segformer} & $\bm{\phi}_{\mathrm{FI}}$ & 22.145 & 0.296  \\

        % U-Net with CBAM & $\bm{\phi} = \mathbf{x} \oplus D(\hat{\mathbf{z}}_T) \oplus D(\hat{\mathbf{z}}_0) \oplus |\mathbf{x} - D(\hat{\mathbf{z}}_0)|$ & $\num{3.288e-3}$ & 24.831 & 0.875 \\

        \midrule

        X-Edit (ours) &  $\bm{\phi}_{\mathrm{FI}}$ & \textbf{24.946} & \underline{0.945} \\
        
        X-Edit (ours) & $\bm{\phi}$ & 24.831 & 0.875 \\

        % U-Net with CBAM (finetuned) & $\bm{\phi} = \mathbf{x} \oplus D(\hat{\mathbf{z}}_T) \oplus D(\hat{\mathbf{z}}_0) \oplus |\mathbf{x} - D(\hat{\mathbf{z}}_0)|$ & \num{4.741e-3} & 24.270 & 0.954 \\
        X-Edit + finetuning (ours) & $\bm{\phi}_{\mathrm{FI}}$ & \underline{24.926} & 0.943 \\ 
        X-Edit + finetuning (ours) & $\bm{\phi}$ & 24.270 & \textbf{0.954} \\
        
        \bottomrule
    \end{tabular}
    }
    \caption{\textbf{Performance metrics for various models and input types.} X-Edit method overperforms baselines both in terms of PSNR and SSIM. Best result in bold, second best underlined.}
    \label{tab:metrics_res_2}
\end{table}

\paragraph{Model Architecture} 
Our primary model for the segmentation task is a U-Net architecture augmented with Convolutional Block Attention Modules (CBAM) \cite{woo2018cbam}. The input dimension is $512 \times 512$. The U-Net is composed by 4 downsampling layers and 4 upsampling layers with skip connections. CBAM enhances feature representation by sequentially applying attention mechanisms along both the channel and spatial dimensions within each convolutional block. Specifically, each CBAM block consists of a Channel Attention Module and a Spatial Attention Module.
Channel Attention Module emphasizes informative feature channels by computing attention weights across channels using global pooling operations, which are then applied to the feature map.
Spatial Attention Module highlights important spatial regions by computing attention weights across spatial dimensions using pooling operations along the channel axis, and applies these weights to the feature map.
By integrating CBAM into the U-Net architecture, we improved the model's ability to focus on relevant features both across channels and spatially, enhancing segmentation performance.

\paragraph{Comparison Models} To the best of our knowledge, no state-of-the-art method is available for the task of detecting and localizing TGIE. We therefore evaluate the effectiveness of our approach by comparing our method against several other baselines. First, we consider a standard U-Net architecture with the same dimensionality as the primary model, but without CBAM modules.
% \nb{The}
We also consider three ViT based architectures. A simple ViT-B~\cite{dosovitskiy2020image} (input $224\times224$) trained from scratch, the SegFormer model \cite{xie2021segformer} (input $512\times512$) and Segment Anything Model (SAM) \cite{kirillov2023segment} (input $512\times512$). SegFormer combines global context capture with efficient multi-scale feature fusion, while SAM \cite{kirillov2023segment} is a state-of-the-art segmentation model known for its ability to segment objects in images with high accuracy.

\paragraph{Inference and Evaluation Metrics} At inference time, given a query image $\mathbf{x}$, we generate a caption using \mbox{BLIP-3~\cite{xue2024xgen}} and we compute the feature $\bm{\phi}$ as described in \cref{eq:phi}. Then we feed $\bm{\phi}$ to a model and obtain a mask estimation $\hat{\mathbf{y}}$. 
% \TODO{For the sake of the experiments, we also define here $\bm{\phi}_{\mathrm{FI}}$ as the input feature defined in FakeInversion paper~\cite{cazenavette2024fakeinversion}. It differs from $\bm{\phi}$ for having grayscale input image and not having the residuals  $|\mathbf{x} - D(\hat{\mathbf{z}}_0)|$ concatenated.} 
We evaluate the models using key standard metrics for reconstruction tasks: Peak Signal-to-Noise Ratio (PSNR) and Structural Similarity Index Measure (SSIM) between the predicted mask $\hat{\mathbf{y}}$ and the ground-truth mask $\mathbf{y}$. The former gives us an indication of how faithful the reconstruction task is on a pixel-level, the latter is useful for evaluating the perceived reconstruction quality.

\begin{table}[t]
    \centering
    % \resizebox{\columnwidth}{!}{%
    \begin{tabular}{cl c c}
        \toprule
         & \textbf{Input} &  \textbf{PSNR} & \textbf{SSIM} \\
        \midrule
        
        (A) & $\mathbf{x}$ & 24.632 & 0.859 \\

        (B) &  $\mathbf{x} \oplus D(\hat{\mathbf{z}}_T)$ & 24.535 & 0.837 \\
        
        (C) &$\mathbf{x} \oplus D(\hat{\mathbf{z}}_0)$ & 24.677 & 0.894 \\

        (D) & $\mathbf{x} \oplus D(\hat{\mathbf{z}}_T) \oplus D(\hat{\mathbf{z}}_0)$ & 24.709 &  0.889 \\
        
        (E) & $\mathbf{x} \oplus D(\hat{\mathbf{z}}_T) \oplus |\mathbf{x} - D(\hat{\mathbf{z}}_0)|$ & 24.576 & 0.893 \\

        \midrule
        &$\bm{\phi}$ & 24.831 & 0.875 \\
        \bottomrule
    \end{tabular}
    % }
    \caption{\textbf{Ablation on input composition.} Performance metrics for X-Edit trained on different combination of partial $\bm{\phi}$.}
    \label{tab:ablation_res}
\end{table}

% \paragraph{Evaluation Metrics} The models were evaluated using three key metrics. The Mean Squared Error (MSE) was used to measure the average squared difference between predicted masks and the ground truth. The Structural Similarity Index Measure (SSIM) was employed to assess the perceptual similarity between predicted and ground truth masks, capturing how visually consistent the predicted mask was with the actual edited regions. Finally, the Peak Signal-to-Noise Ratio (PSNR) evaluated the ratio between the maximum possible signal power and the power of corrupting noise, providing an overall measure of prediction quality.

\begin{figure*}[t] 
    \centering
    \includegraphics[width=\textwidth]{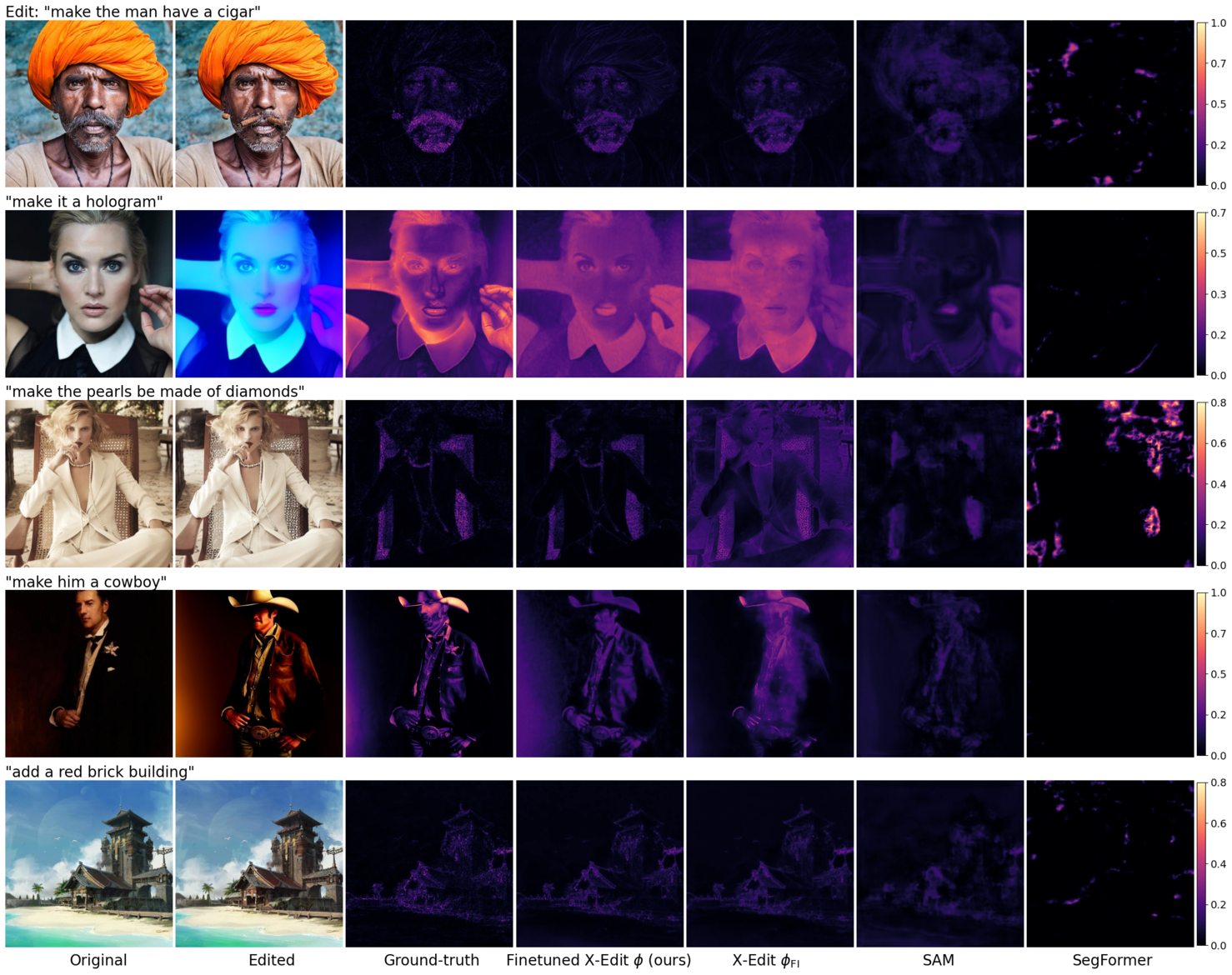} 
    \caption{
\textbf{Comparison of predicted masks for edited images.} 
From left to right: original image, edited image, ground truth mask indicating the edited regions, predicted mask from X-Edit finetuned on $\bm{\phi}$, X-Edit on $\bm{\phi}_{\textrm{FI}}$, SAM and SegFormer. X-Edit finetuned on $\bm{\phi}$ (4th column) outperforms other models by more accurately capturing both the shape and placement of edits, demonstrating finer boundary alignment and better preservation of details in complex regions. This improvement highlights X-Edit's effectiveness in maintaining contextual coherence and producing higher-fidelity masks for intricate modifications. 
% The original model uses composite input features defined as $\bm{\phi} = \mathbf{x} \oplus D(\hat{\mathbf{z}}_T) \oplus D(\hat{\mathbf{z}}_0) \oplus |\mathbf{x} - D(\hat{\mathbf{z}}_0)|$. The fine-tuned model, initialized from the converged  model, trained with specific loss weightings: $\lambda_{\text{flat}} = 0.1$, $\lambda_{\text{edge}} = 3.0$, and $\lambda_{\text{R}} = 0.5$.
}
    \label{fig:comparison}
\end{figure*}

\paragraph{Optimization and Finetuning} For optimization, we utilized the AdamW optimizer \cite{loshchilov2017fixing}, with a learning rate of \num{1e-4} and a weight decay of \num{1e-3}. To manage the learning rate schedule, we used a cosine annealing warm restarts scheduler \cite{loshchilov2016sgdr}, which adjusted the learning rate dynamically during training. The hyperparameters $\lambda_{\text{flat}}$, $\lambda_{\text{edge}}$, $\lambda_{\text{R}}$, and $\alpha$ were empirically determined to balance the contribution of each loss component: $\lambda_{\text{flat}} = 0.1$, $\lambda_{\text{edge}} = 3.0$, and $\lambda_{\text{R}} = \lambda_{\text{S}} = 0.5$, $\alpha=0.2$. Note that we impose the constraint $\lambda_{\text{R}} + \lambda_{\text{S}} = 1$ to simplify the tuning process and ensure a balanced trade-off between the relevance and segmentation losses.  
\section{Results}
\label{sec_results}

\begin{figure}[t]
    \centering
    \includegraphics[width=\columnwidth]{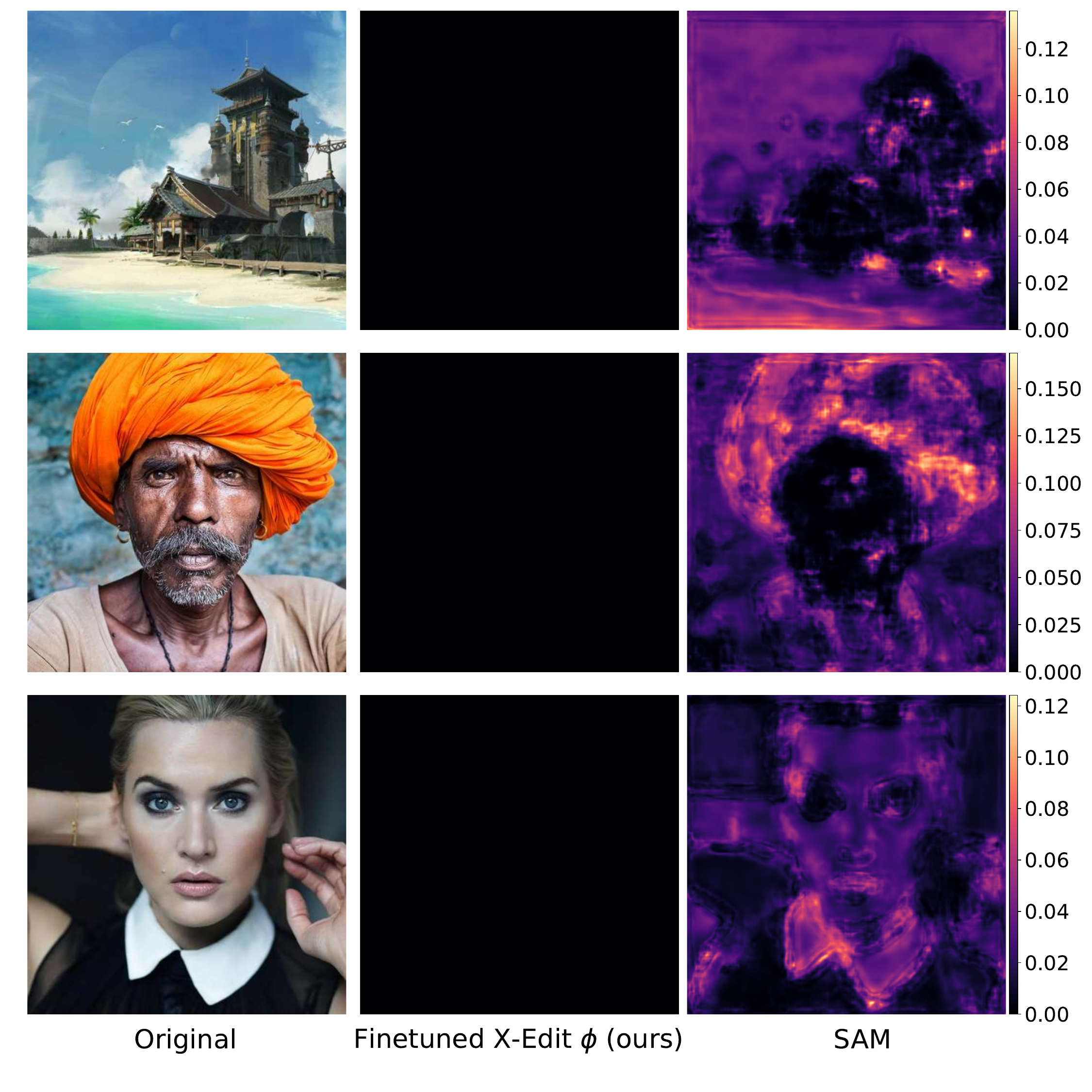}
    \caption{\textbf{Comparison of predicted mask for original images.} 
    X-Edit manages to predict almost blank masks for original images, while SAM tends to produce false positives.}
    \label{fig:original}
\end{figure}

\begin{figure}[t]
    \centering
    \includegraphics[width=\columnwidth]{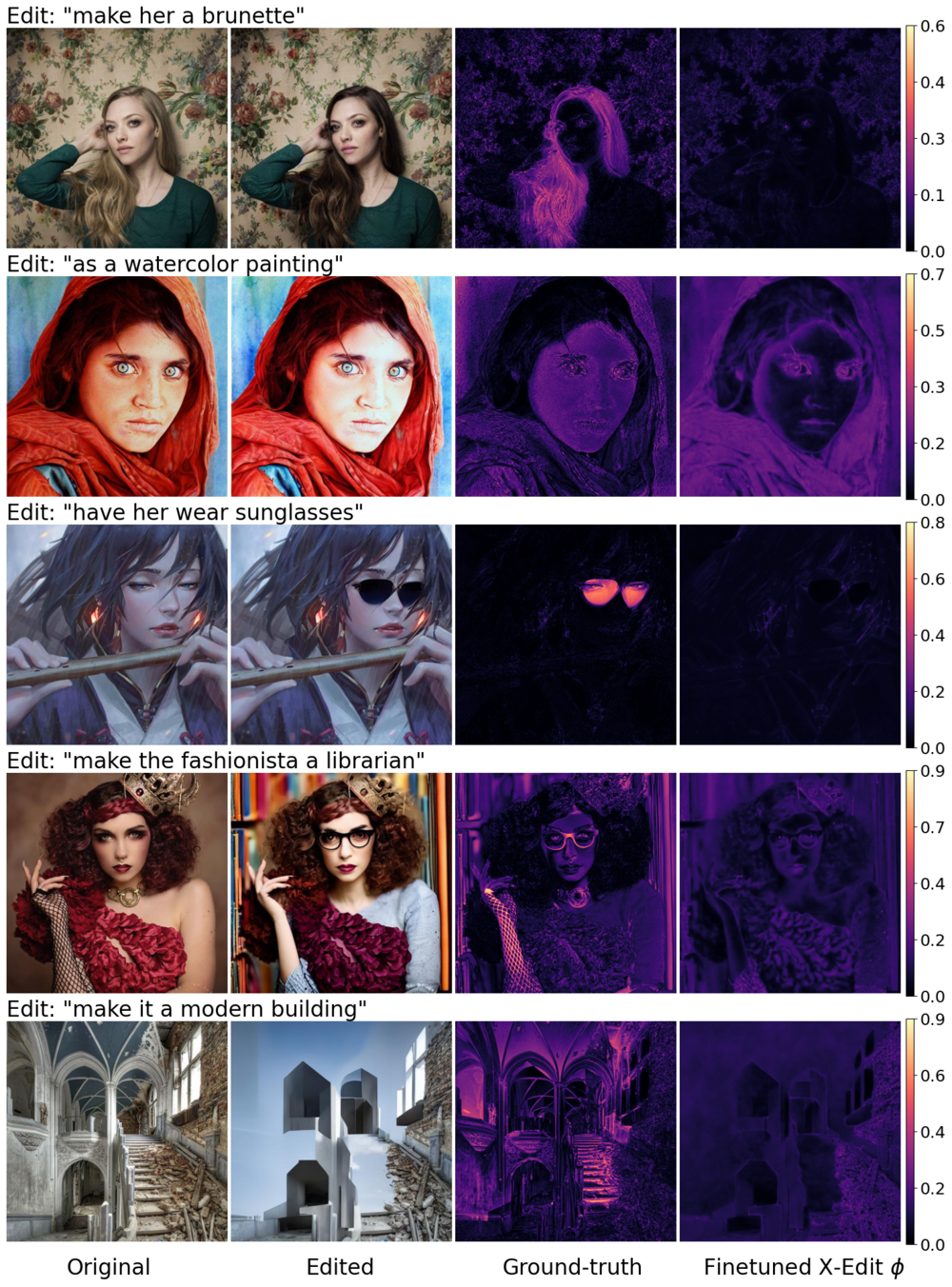}  
    \caption{\textbf{Failure cases}. These example highlight challenges with color changes, style shifts, and occlusions that require both structural adaptation and color inference.}
    \label{fig:failed_cases}
\end{figure}

\vspace{.4em}\noindent\textbf{Quantitative results.} \Cref{tab:metrics_res_2} shows quantitative results for our proposed method with respect to the baselines and different network inputs. 
We first build two preliminary baselines consisting in SAM and a U-Net directly trained on the RGB image $\mathbf{x}$. We notice how U-Net outperforms the state-of-the-art segmentation model, expecially in SSIM.
We therefore keep the U-Net and we train it on same feature used in FakeInversion~\cite{cazenavette2024fakeinversion}, $\bm{\phi}_{\mathrm{FI}}$, comparing with SegFormer and ViT. U-Net is the best performing of the three both in PSNR and SSIM, and we notice a boost with respect to the same model trained on RGB input $\mathbf{x}$. We then train the X-Edit architecture both on $\bm{\phi}_{\mathrm{FI}}$ and  $\bm{\phi}$ and we notice a substantial improvement in PSNR and SSIM for the former. X-Edit trained on $\bm{\phi}_{\mathrm{FI}}$ shows the best PSNR of all the experiments, but we achieve the best SSIM only with the finetuning of X-Edit on the complete feature $\bm{\phi}$. To explain this, we note how $\bm{\phi}$ is an extension of $\bm{\phi}_{\mathrm{FI}}$, and the finetuning procedure is designed to discourage the network to excessively focus on high-frequency regions. As a result, X-Edit finetuned on $\bm{\phi}$ produces prediction maps that are perceptually more similar to the ground-truth, albeit less faithful on a pure pixel level. 
%
% We can notice how using the complete feature $\bm{\phi}$ outperforms the baseline using the input image $\mathbf{x}$. In particular, the non-finetuned version achieves the best PSNR, while the finetuned version achieves the best SSIM. This can be explained by the adoption of the relevance loss $\mathcal{L}_\mathrm{R}$ (\cref{eq:relevance_loss}), that discourages the network to excessively focus on high-frequency regions. As a result, the network produces prediction maps that are perceptually more similar to the ground-truth, albeit less faithful on a pure pixel level. 

\vspace{.4em}\noindent\textbf{Qualitative results.} We display qualitative results for edited images in \cref{fig:comparison}. We can see how both X-Edits effectively highlights edges in edited images, as it captures areas where the reconstruction deviates from the original image, which are often located at the transition zones of edits. Our method enhances the sensitivity to edit-related edges, supporting accurate edit localization and boundary detection. The finetuning adjustments further optimize this sensitivity, allowing the model to refine its focus on edited regions while minimizing background noise in unedited areas. We also show examples of prediction of original images in \cref{fig:original}, to highlight how our model is capable to keep false positives at bay. The finetuned X-Edit outputs a blank estimation, while SAM tends to produce some estimation even if the image is unedited. This shows us the importance of SSIM as a metric to evaluate predictions: in \cref{tab:metrics_res_2} SAM has a better PSNR than the baseline U-Net, but the very low SSIM leads to an excess of false positives. We are also providing qualitative results for a smaller set of out-of-distribution data and TGIE methods (FPE~\cite{liu2024towards}, MasaCtrl~\cite{cao2023masactrl}, PnP~\cite{tumanyan2023plug}) in the supplementary material.

\vspace{.4em}\noindent\textbf{Failure cases.} While our model generally performs well in distinguishing original from edited images and effectively captures edge-like features in edited regions, it still faces challenges with certain types of edits, as shown in \cref{fig:failed_cases}. The addition of the relevance loss $\mathcal{L}_\mathrm{R}$ improves overall performance but can introduce issues when edits involve high-frequency areas, possibly due to susceptibility in the loss function that may require further hyperparameter tuning. We also observe a tendency to fail in localizing edits involving the addition of dark, flat objects that occlude original features (\eg, sunglasses, mustaches), as the model needs to infer hidden content rather than recognizing an object with altered color. Additionally, edits that are too subtle or involve low-entropy changes (like in row 5 of \cref{fig:failed_cases}) may not provide enough information for the model to detect the alterations.

% Finally, Color change is a more subtle alteration because the object's form and structure remain visible, but occlusion removes or hides part of the image, presenting a more significant challenge for our model. The model needs to infer what’s behind the occluded area, whereas for a color change, it simply needs to recognize that the object is the same, just with a different color.

\vspace{.4em}\noindent\textbf{Ablation.}
In \cref{eq:phi}, we defined our network input $\bm{\phi}$ as the concatenation of the RGB image, the decoded noise map, the reconstructed image and the image residual after the DDIM inversion. In this section we perform an ablation study over the input composition. \cref{tab:ablation_res} shows PSNR and SSIM for X-Edit fed with various concatenations of partial $\bm{\phi}$ we name (A)-(E). We notice that configuration (A) performs better than (B) despite having less information to exploit, but the addition of $D(\hat{\mathbf{z}}_0)$ is beneficial both in terms of PSNR and SSIM. The combination of the two decoded noise map in configuration (D) gives a boost that is unmatched by configuration (E), where the $D(\hat{\mathbf{z}}_0)$ is replaced by the residual. The complete feature $\bm{\phi}$ shows better fidelity to the ground-truth according to the PSNR, albeit being less perceptually similar (lower SSIM) than configuration (C), (D) and (E). 
% \TODO{Maybe we should add something to avoid the question ``why did you go with $\bm{\phi}$ then''}. 
Despite the slightly lower SSIM, we choose the complete feature $\bm{\phi}$ because it provides better overall fidelity to the ground truth as indicated by the higher PSNR. Additionally, incorporating all components into $\bm{\phi}$ allows the model to capture a more comprehensive range of discrepancies introduced by edits.

\section{Conclusions}
\label{sec_conclusion}
In this paper, we introduce X-Edit, a method for localizing text-guided image editing, and a dataset composed by paired original end edited image to fill the lack of resources in the field. To our knowledge, X-Edit is the first method specifically addressing localization of text-guided image edits.
% Our method outperforms several baselines in terms of PSNR and SSIM and is able to keep false positives at bay. 
Experimental results demonstrate that X-Edit outperforms several baselines, including state-of-the-art segmentation models, in terms of PSNR and SSIM metrics. The qualitative evaluations show that X-Edit can accurately highlight the specific areas where edits have occurred, while maintaining low false positive rates on unedited images. This represents a significant advancement in developing robust forensic tools capable of detecting and explaining manipulations introduced by advanced image editing techniques.
% Future work will be devoted to explore a better finetuning strategy for the composite loss aimed to find a better balance for high- and low-frequency frequency hyperparameters \nb{to enhance performance on challenging edits}. 
In the future, we plan to extend our approach to handle a wider variety of editing techniques and to improve the model's ability to detect edits in flat regions or when high-frequency cues are minimal.

% 1. Does well in terms of detecting original vs. edited images. 

% 2. Does well to detect edges-looking masks for edited images (not for original images) =``Edge'' detector for edited images - could be related to the dataset, as some edits are not meaningful or too subtle, and for high entropy images (edit caption:``make it a photo''), yet when going through the diffusion-based editing pipeline like InstructPix2Pix, there are discrepancies in edges.

% 3. Images that failed - not enough information, connect to entropy and discussion of the lack of features.

% 4. Color change is a more subtle alteration because the object's form and structure remain visible, but occlusion removes or hides part of the image, presenting a more significant challenge for our model. The model needs to infer what’s behind the occluded area, whereas for a color change, it simply needs to recognize that the object is the same, just with a different color.

% 5. Still needs improvement for flat regions 
{
    \small
    \bibliographystyle{ieeenat_fullname}
    \bibliography{references}
}
\clearpage
\setcounter{page}{1}
\maketitlesupplementary

\renewcommand{\thesection}{\Alph{section}}

\section{Overview}
This document is structured as follows:
\begin{itemize}
    \item \cref{sec:supp_quali}: Additional qualitative results;
    \item \cref{sec:supp_histograms}: Prediction distributions for original images;
    \item \cref{sec:supp_ood}: Qualitative results on out-of-distribution dataset.
\end{itemize}

\section{Additional Qualitative results}
\label{sec:supp_quali}
\begin{figure*}[t] 
    \centering
    \includegraphics[width=\textwidth]{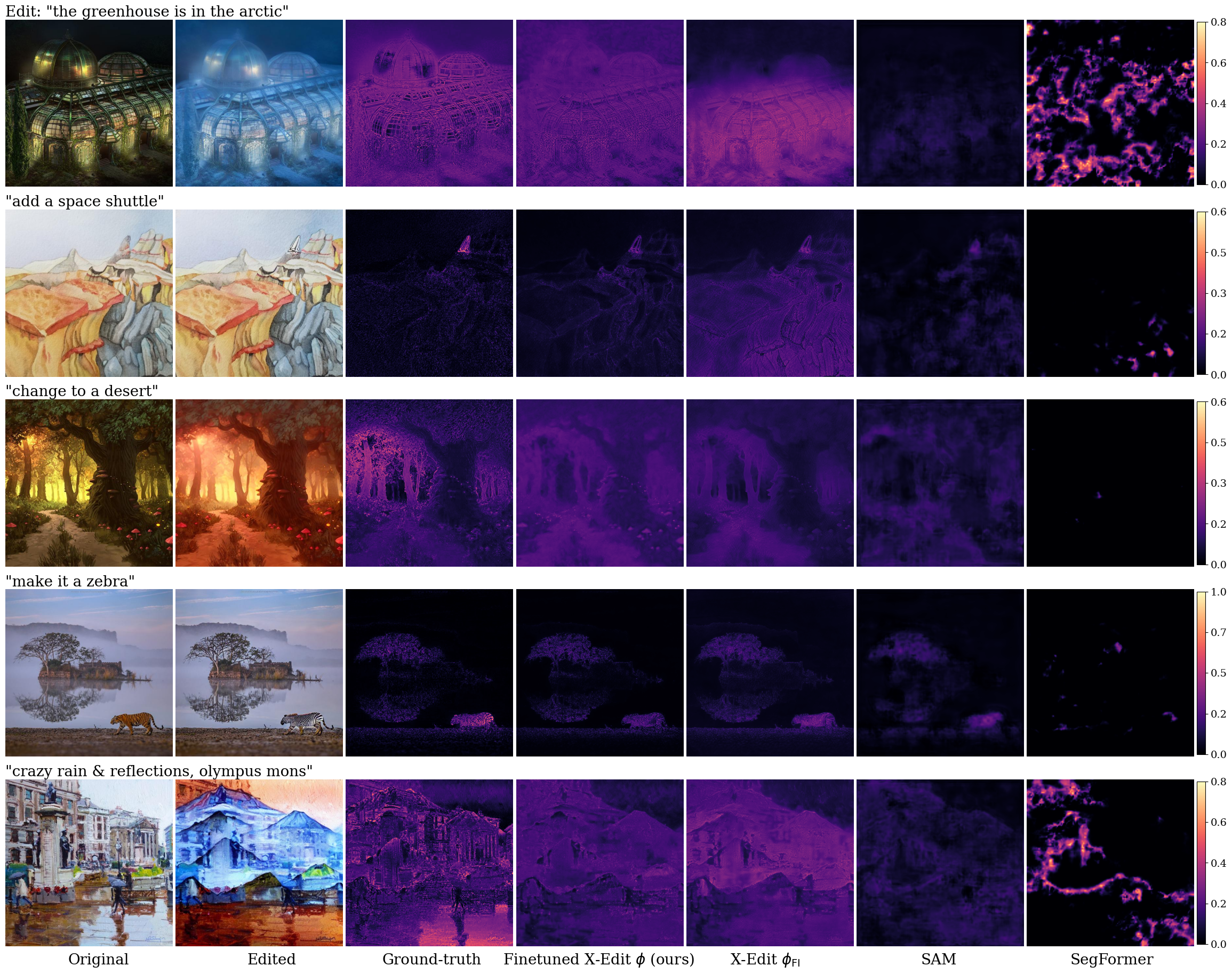} 
    \caption{
\textbf{Additional comparison of predicted masks for edited images.} 
From left to right: original image, edited image, ground truth mask indicating the edited regions, predicted mask from X-Edit finetuned on $\bm{\phi}$, X-Edit on $\bm{\phi}_{\textrm{FI}}$, SAM and SegFormer. X-Edit finetuned on $\bm{\phi}$ (4th column) outperforms other models by more accurately capturing both the shape and placement of edits, demonstrating finer boundary alignment and better preservation of details in complex regions. This improvement highlights X-Edit's effectiveness in maintaining contextual coherence and producing higher-fidelity masks for intricate modifications.  
}
    \label{fig:comparison_2}
\end{figure*}

In this section we complement \cref{fig:comparison} by showing in \cref{fig:comparison_2} some more qualitative results we could not include in the main paper for space constraints.

\section{Prediction distributions for original images}
\label{sec:supp_histograms}

\begin{figure}[t]
    \centering
    \includegraphics[width=\columnwidth]{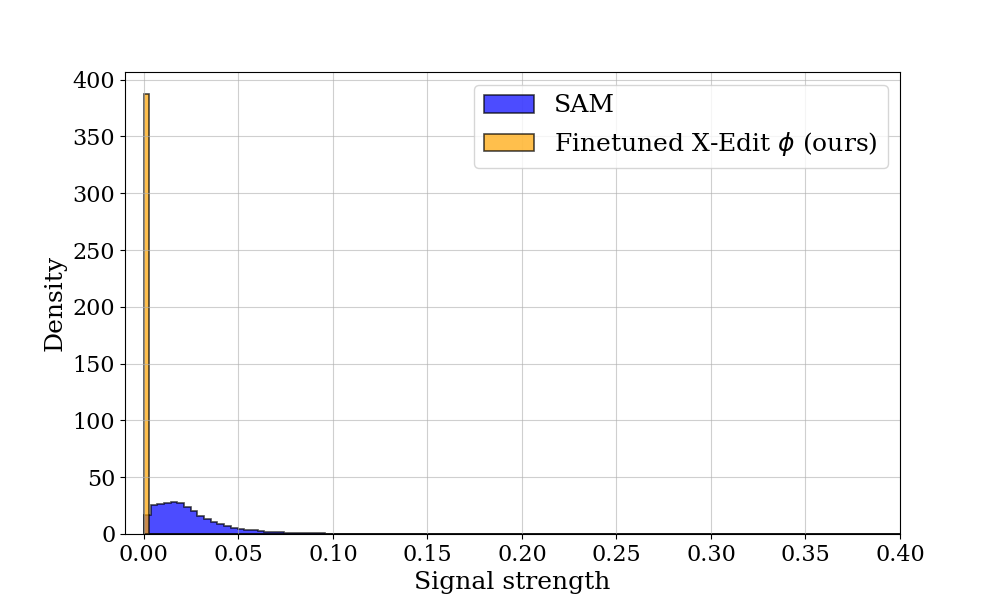}
    \caption{\textbf{Comparison of model distributions for original images.} The histograms display the density distributions of predicted mask values for SAM and Finetuned X-Edit $\phi$ models.}
    \label{fig:histograms}
\end{figure}

In this section, we analyze the distribution of predicted masks generated on original images by two segmentation models: SAM~\cite{kirillov2023segment} and our finetuned X-Edit method. We already shown in \cref{fig:original} how our model is better than SAM at predicting blank mask for original images. In \cref{fig:histograms} we provide a broader visualization by plotting prediction values over the entire test dataset. We can observe how X-Edit's histogram is concentrated around $0$, while SAM's shows a significant spread that translates in non-zero prediction mask displayed in \cref{fig:original}.

% For each model, we process the test images and generate predicted masks for both the original and edited versions of each image. The predicted masks are flattened and concatenated to create comprehensive datasets for analysis.

% These plots display the density of predicted mask values, allowing us to compare the distributions between SAM and the finetuned X-Edit method. We create two subplots: one for the original images and one for the edited images. This comparison highlights the models' sensitivity to image edits and their ability to generalize to altered inputs.

\section{Qualitative results on out-of-distribution dataset}
\label{sec:supp_ood}
In this section, we provide some additional qualitative results on an out-of-distribution dataset we build.

\subsection{Out-of-distibution dataset}
We build the original part of our out-of-distribution by randomly selecting $100$ images from the Flickr30k~\cite{young2014image} test split using the Hugging Face Datasets library~\cite{lhoest-etal-2021-datasets}. We choose each image based on a minimum dimension criterion of $500$ pixels to ensure high-quality inputs suitable for editing. Alongside the images, we extract the first caption provided for each image to use as reference text.
Besides InstructPix2Pix~\cite{instructpix2pix_2023}, we select three additional state-of-the-art TGIE methods for editing images:
\begin{enumerate}
    \item FPE~\cite{liu2024towards} uses self-attention control to guide the diffusion process towards the target prompt
    \item MasaCtrl~\cite{cao2023masactrl} editing allows to set up the mutual self-attention controller with specified steps and layers, thus registering the attention editor within the diffusion pipeline.
    \item PnP~\cite{tumanyan2023plug} manipulates internal spatial features and self-attention components during the diffusion process.
\end{enumerate}
By applying these editing methods, we generate two edited versions for each image in the test set. These edited images are used to assess the performance of our models in handling out-of-distribution data and to evaluate their robustness to different types of image alterations.

\subsection{Results}
\begin{figure*}[t] 
    \centering
    \includegraphics[width=0.5\textwidth]{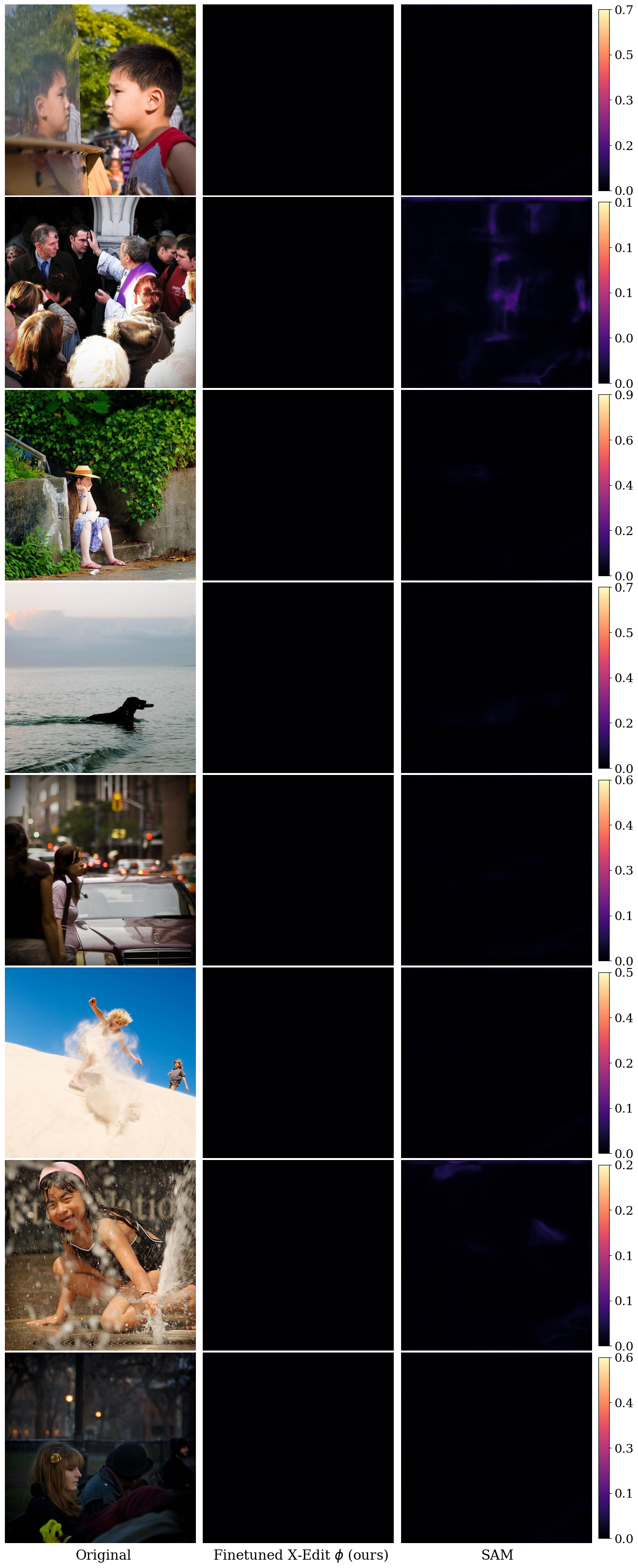} 
    \caption{
\textbf{Comparison of predicted masks for original images on out-of-distribution flickr30k images.} 
% X-Edit manages to predict almost blank masks for original images, while SAM tends to produce false positives.
}
    \label{fig:suppl_OOD_original_masks}
\end{figure*}

\cref{fig:suppl_OOD_original_masks} shows some examples of our X-Edit method vs SAM on Flickr30k original images. We notice how our method still produces blank predictions while SAM produces some non-zero maps in some cases.

\begin{figure*}[t] 
    \centering
    \includegraphics[width=0.9\textwidth]{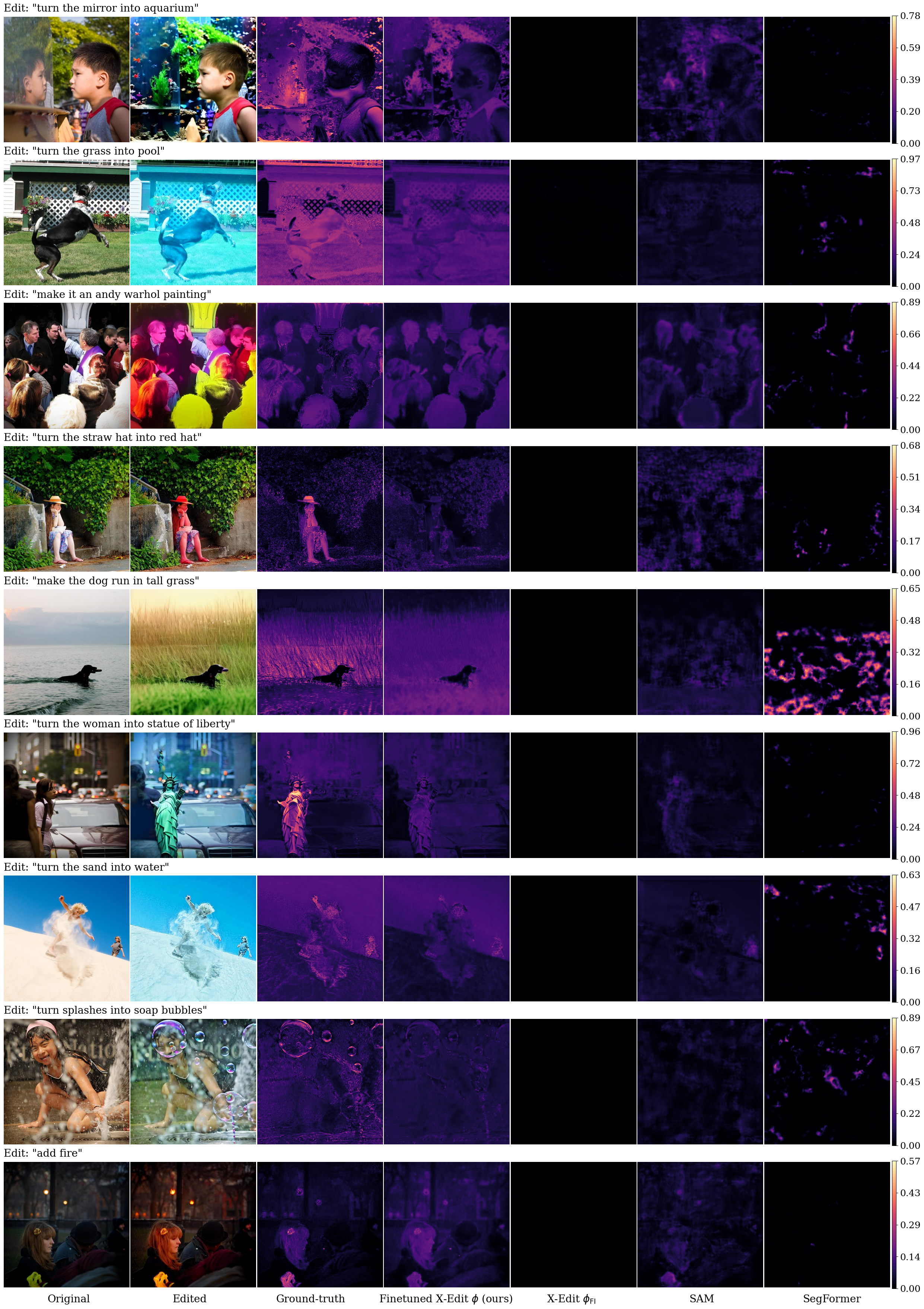} 
    \caption{
\textbf{Comparison of predicted masks for edited images with InstructPix2Pix method on out-of-distribution Flickr30k images.}
}
    \label{fig:suppl_OOD_InstPix2Pix}
\end{figure*}

\begin{figure*}[t] 
    \centering
    \includegraphics[width=0.9\textwidth]{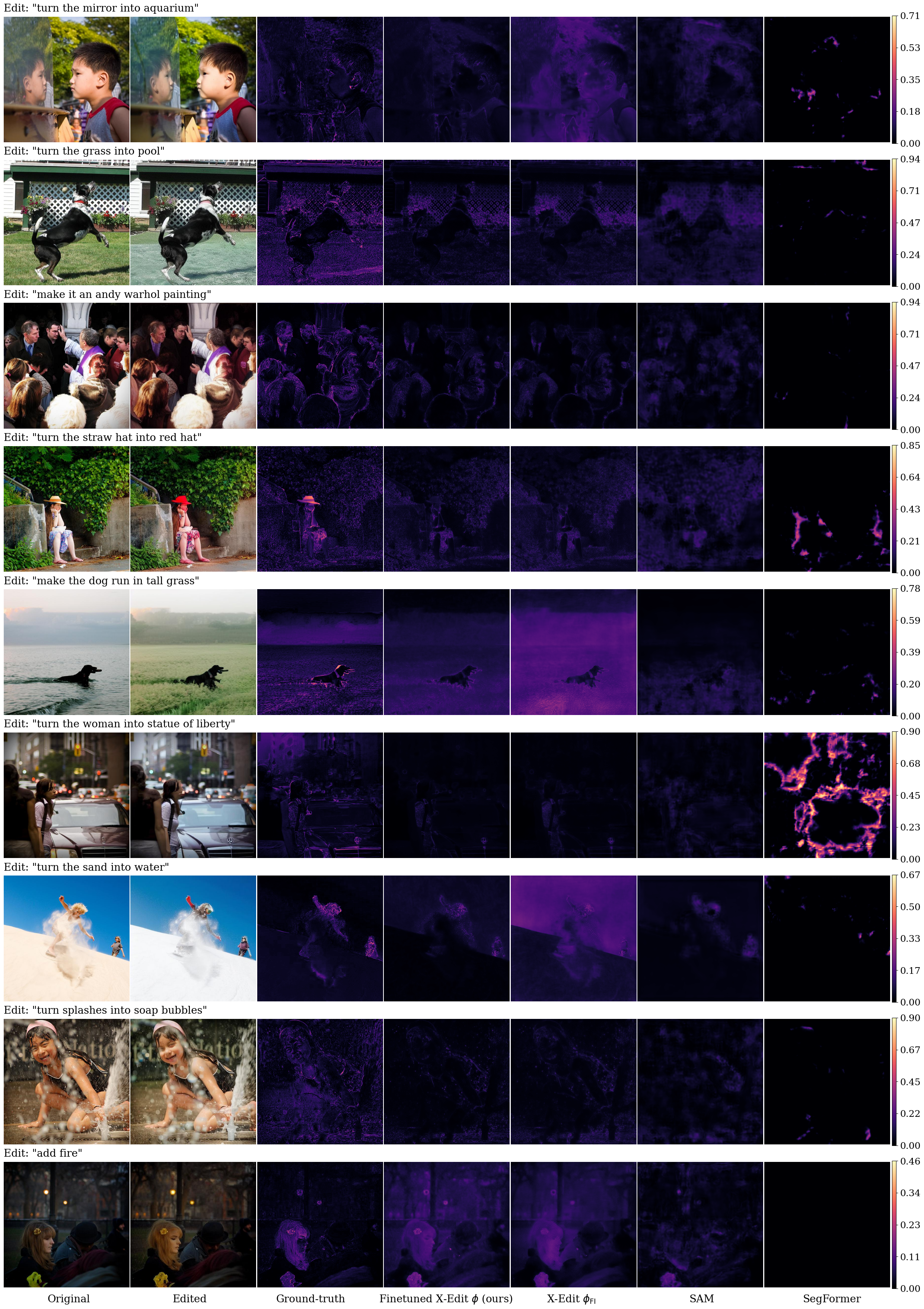} 
    \caption{
\textbf{Comparison of predicted masks for edited images with out-of-distribution FPE method on out-of-distribution Flickr30k images.}
}
    \label{fig:suppl_OOD_FPE}
\end{figure*}

\begin{figure*}[t] 
    \centering
    \includegraphics[width=0.9\textwidth]{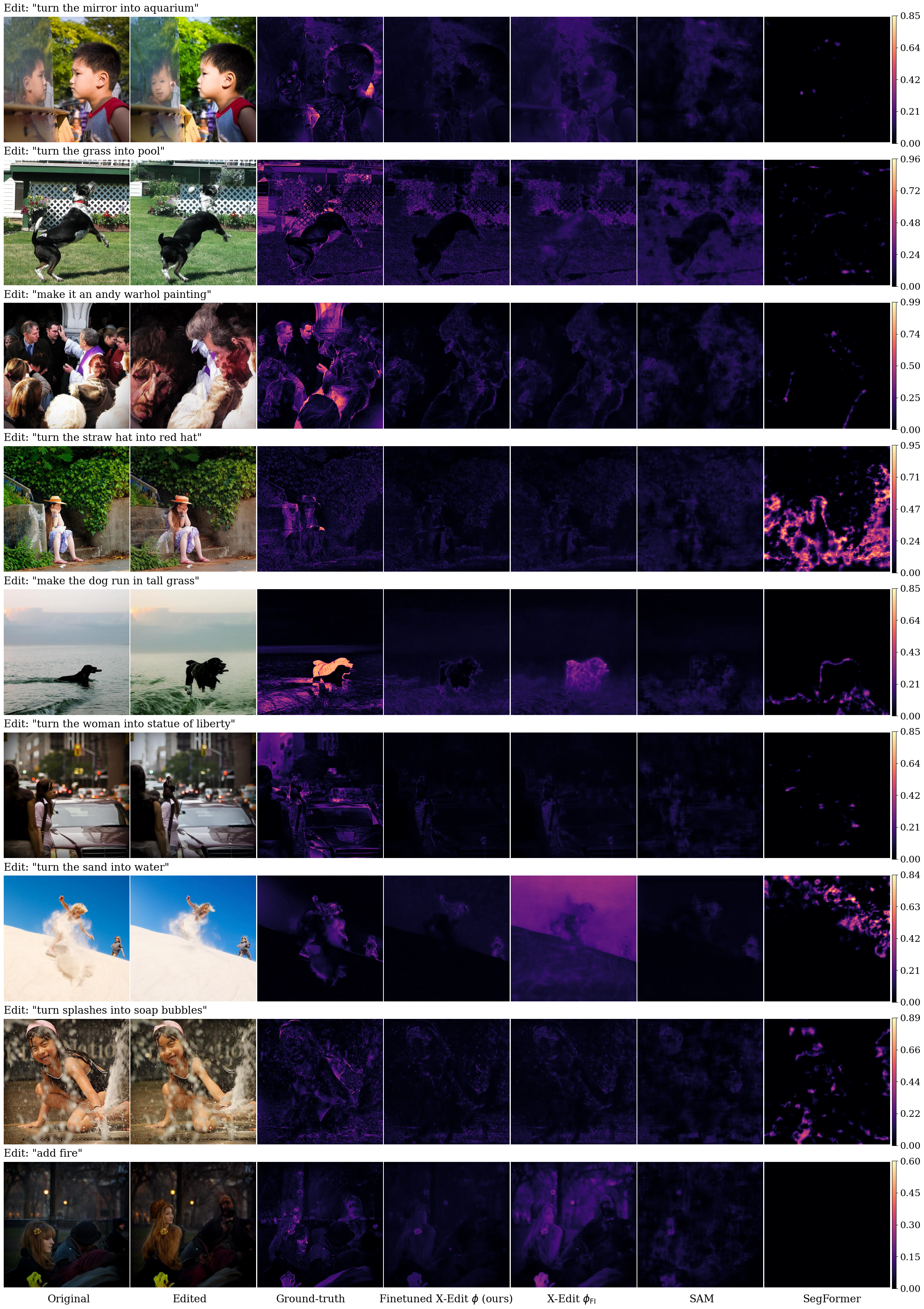} 
    \caption{
\textbf{Comparison of predicted masks for edited images with MasaCtrl method on out-of-distribution Flickr30k images.}
}
    \label{fig:suppl_OOD_MasaCtrl}
\end{figure*}

\begin{figure*}[t] 
    \centering
    \includegraphics[width=0.9\textwidth]{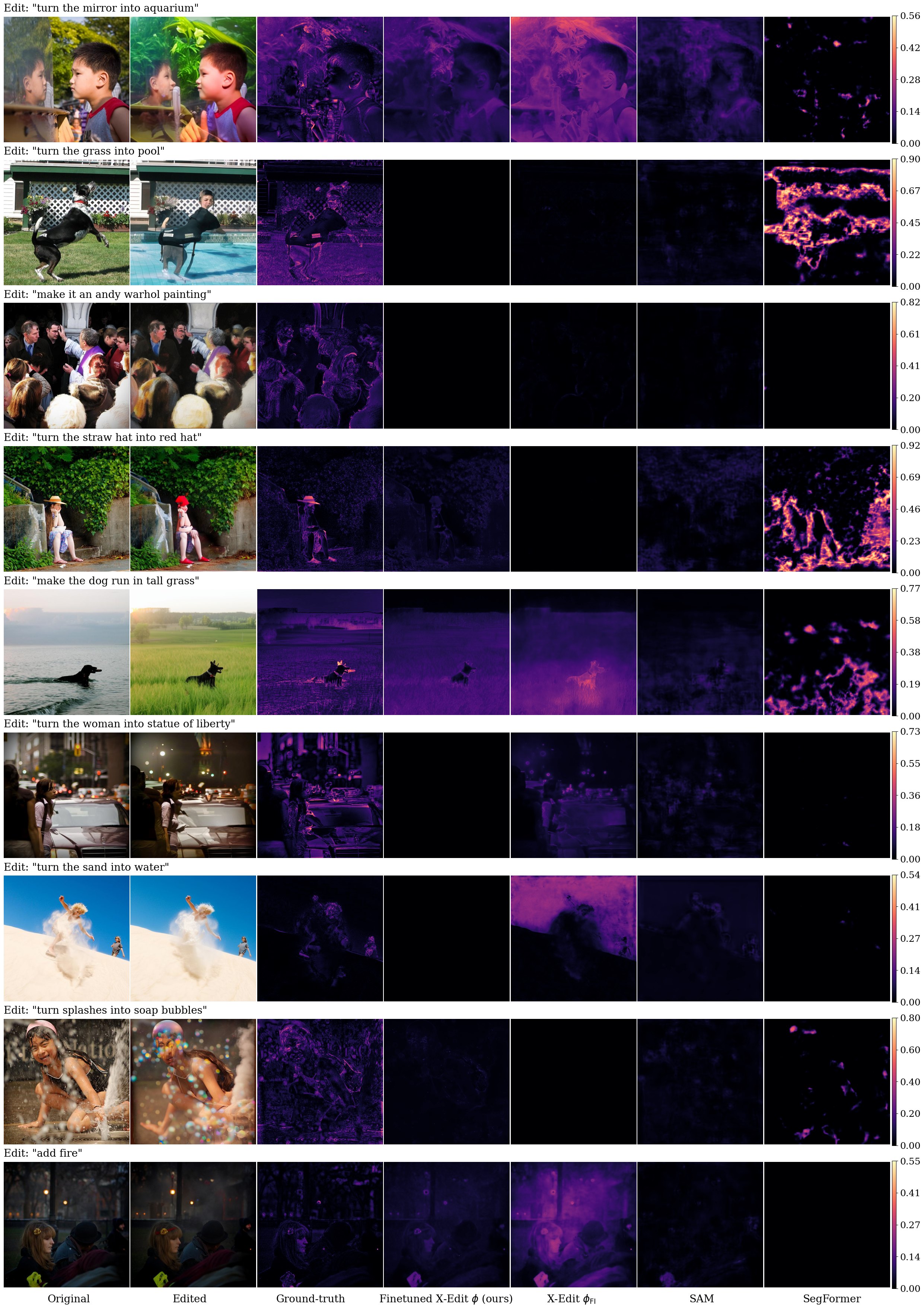} 
    \caption{
\textbf{Comparison of predicted masks for edited images with PnP method on out-of-distribution Flickr30k images.}
}
    \label{fig:suppl_OOD_PnP}
\end{figure*}

\cref{fig:suppl_OOD_InstPix2Pix}, \cref{fig:suppl_OOD_FPE}, \cref{fig:suppl_OOD_MasaCtrl} and \cref{fig:suppl_OOD_PnP} show some examples of our X-Edit method vs SAM~\cite{kirillov2023segment} and SegFormer~\cite{xie2021segformer} on edited images generated by InstructPix2Pix, FPE, MasaCtrl and PnP respectively. We keep the same original images across the figures to enable a clearer comparison, not just between the segmentation methods, but also across the variety of editing outputs with its unique challenges. We can observe how for InstructPix2Pix and FPE our finetuned X-Edit produces masks that are very close to the ground-truth, while SAM tends to produce shallower masks and SegFormer is prone to false positives. This shows X-Edit's strength in precisely identifying edited regions, especially for localized changes like ``add fire" or ``turn the sand into water" where segmentation closely matches edits. Conversely, SAM's shallow masks fail to capture precise boundaries, and SegFormer's false positives indicate misclassification of unrelated regions. Regarding MasaCtrl and PnP edits, we notice how X-Edit on $\bm{\phi}_{\textrm{FI}}$ seems to be best model, with SAM and SegFormer showing same limitation as before. 
Edits by MasaCtrl and PnP, involving complex, large-scale changes like pose or style shifts, present additional challenges. X-Edit on $\bm{\phi}_{\textrm{FI}}$ effectively identifies the broader impacted regions while maintaining coherence. However, SAM and SegFormer struggle, underestimating edit scopes or misapplying changes to unrelated areas.

We selected edited images to showcase diverse challenges, evaluating how X-Edit and other models predict edited regions. Examples include additive edits (\eg, ``add fire"), background or foreground modifications (\eg, ``turn the grass into pool"), stylistic changes (\eg, ``make it an Andy Warhol painting"), and targeted edits (\eg, ``turn the straw hat into a red hat"), each testing specific capabilities like seamless integration, spatial coherence, style adaptation, and precise localization. These scenarios are crucial for testing models' ability to accurately identify changed regions. The range of chosen edits enables the assessment of models' robustness in capturing edits while maintaining accuracy. Importantly, the examples highlight both successes and failures. For instance, FPE identifies edited regions well in context-aware tasks but struggles with stylistic transformations, introducing artifacts and failing to preserve spatial coherence. Similarly, MasaCtrl excels in structural edits and pose adjustments but can distort facial features or introduce unintended changes, especially in close-ups. For example, edits involving faces or abstract changes sometimes show exaggerated or inaccurate deformations.

% WARNING: do not forget to delete the supplementary pages from your submission 
% \input{sec/X_suppl}

\end{document}